\documentclass[final,5p,times,twocolumn]{elsarticle}

\usepackage{amsmath,amssymb,amsfonts}
\usepackage{algorithm, algorithmic}
\usepackage{multirow, url}

\journal{Neurocomputing}

\begin{document}
	
	\begin{frontmatter}
		\title{Conceptual Text Region Network: Cognition-Inspired Accurate Scene Text Detection}
		
		\author{Chenwei~Cui\fnref{label1}}
		\author{Liangfu~Lu\corref{cor1}\fnref{label1}}
		\author{Zhiyuan~Tan\fnref{label2}}
		\author{Amir~Hussain\fnref{label2}}
		
		\address[label1]{School of Mathematics, Tianjin University, Tianjin, 300350, China}
		\address[label2]{School of Computing, Merchiston Campus, Edinburgh Napier University, Edinburgh EH10 5DT, Scotland, U.K.}
		
		\cortext[cor1]{Corresponding Author: \url{liangfulv@tju.edu.cn}}
		
		\begin{abstract}
			Segmentation-based methods are widely used for scene text detection due to their superiority in describing arbitrary-shaped text instances. However, two major problems still exist: 1) current label generation techniques are mostly empirical and lack theoretical support, discouraging elaborate label design; 2) as a result, most methods rely heavily on text kernel segmentation which is unstable and requires deliberate tuning. To address these challenges, we propose a human cognition-inspired framework, termed, Conceptual Text Region Network (CTRNet). The framework utilizes Conceptual Text Regions (CTRs), which is a class of cognition-based tools inheriting good mathematical properties, allowing for sophisticated label design. Another component of CTRNet is an inference pipeline that, with the help of CTRs, completely omits the need for text kernel segmentation. Compared with previous segmentation-based methods, our approach is not only more interpretable but also more accurate. Experimental results show that CTRNet achieves state-of-the-art performance on benchmark CTW1500, Total-Text, MSRA-TD500, and ICDAR 2015 datasets, yielding performance gains of up to 2.0\%. Notably, to the best of our knowledge, CTRNet is among the first detection models to achieve F-measures higher than 85.0\% on all four of the benchmarks, with remarkable consistency and stability.
		\end{abstract}
		
		
	\end{frontmatter}

	\section{Introduction}
	Scene text detection has received much attention from researchers because of its wide applications such as image translation, road sign recognition, and license plate reading. On account of the recent developments in object detection \cite{n1, n2} and semantic segmentation \cite{n3, n4, n5} based on Convolutional Neural Networks (CNNs) \cite{n6, n7}, scene text detection has achieved substantial progress \cite{n8, n9, n10, n11, n12, n13}. Over the past few years, many deep learning-based methods have been proposed to conquer one of its most challenging tasks, arbitrary-shaped text detection. Among these methods, segmentation-based approaches \cite{n10, n11, n12, n13} have become the mainstream because such methods are more capable of describing text instances of irregular shapes. However, two major problems still exist: 1) current label generation techniques adopt rule-based algorithms \cite{n10} or simple operations such as polygon clipping \cite{n11, n13}, which makes them mostly empirical and lacks theoretical support, discouraging elaborate label design; 2) as a result, most methods \cite{n10, n11, n13} rely heavily on the unstable text kernel segmentation, which means directly segmenting the center lines of text instances, to separate adjacent text blocks. Such approach is deficient since it is prone to noise, as stated in \cite{n10}, and requires deliberate tuning, as practiced in \cite{n11}.
	
	To settle these problems, we introduce a human cognition-inspired scene text detection framework called Conceptual Text Region Network (CTRNet) (see Fig.~1). It includes a class of novel label generation tools termed Conceptual Text Regions (CTRs), a Feature Pyramid Network-based (FPN-based) \cite{n14} segmentation network, and a well-designed post-processing algorithm. Observing the solid advantages of introducing cognition-based computing \cite{n15, n16, n17}, we design CTRs to represent the text instances inside our mind when we perceive arbitrary-shaped text instances. Specifically, as shown in Fig.~1~(c), a CTR is a well-defined rectangular area that mathematically inheres a smooth bijection from itself to the corresponding arbitrary-shaped text instance. The very bijectivity allows us to not only define with ease the width, height, and distortion of an arbitrary-shaped text instance but also transfer the geometric features defined within a CTR directly into its corresponding text instance (see Fig.~1). Afterwards, we utilize the segmentation network to learn the geometric features generated by CTRs in order to make predictions on unannotated images. Finally, a robust and interpretable post-processing algorithm makes full use of the geometric features and converts them into the detection result.
	
	\begin{figure*}[ht]
		\begin{center}
			\includegraphics[width=0.9\textwidth]{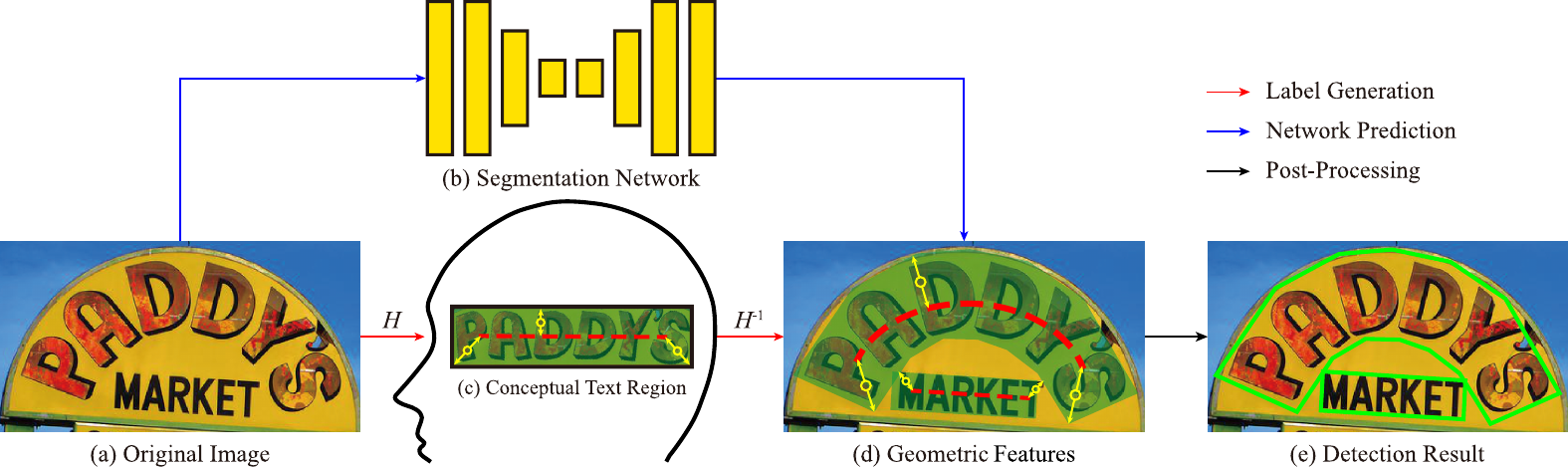}
			\caption{Overview of the CTRNet framework. The red arrows represent the label generation process, during which the original text instances are mapped into their respective CTRs, and the geometric features are generated. The blue arrows, on the other hand, represent the network prediction pipeline, during which the segmentation network substitutes the functionalities of CTRs and predicts the geometric features, given only the original image as input. The black arrow represents the post-processing algorithm which converts the geometric features into the final detection result.}
		\end{center}
	\end{figure*}
	
	To demonstrate the effectiveness of our proposed CTRNet, we conduct extensive experiments on four challenging benchmark datasets including CTW1500 \cite{n18}, Total-Text \cite{n19}, MSRA-TD500 \cite{n20}, and ICDAR 2015 \cite{n21}. Among these datasets, CTW1500 and Total-Text are explicitly designed for arbitrary-shaped text detection. MSRA-TD500 and ICDAR 2015, on the other hand, focus on multi-oriented text. CTRNet consistently outperforms state-of-the-art methods on all four of the datasets in terms of F-measure, yielding performance gains of up to 2.0\%. Notably, to the best of our knowledge, CTRNet is among the first methods to achieve F-measures higher than 85.0\% on all these datasets. 
	
	In summary, our contributions are threefold: 1) We propose CTRs as a class of well-designed label generation tools which inhere good properties such as bijectivity; 2) We propose a segmentation network and a post-processing algorithm that are capable of learning and processing geometric features, omitting the need for text kernel segmentation while establishing an accurate and interpretable inference pipeline. 3) Our CTRNet achieves state-of-the-art performances on a comprehensive set of benchmarks, showing great accuracy and consistency.
	
	\section{Related Work}
	Deep learning-based text detection methods have achieved outstanding results over the past few years. The majority of these methods utilize CNNs and can be roughly categorized into anchor-based methods and segmentation-based methods. In addition, it is of interest that text detection methods in general adopt specific and tailored label generation and post-processing methodologies to utilize the geometric nature of text instances. Thus, we also introduce the label generation and post-processing techniques adopted by text detection methods.
	
	\subsection{Anchor-Based Methods}
	Anchor-based methods are often based on popular object detection frameworks such as Faster R-CNN \cite{n1} and SSD \cite{n2}. 
	
	TextBoxes \cite{n22} modifies anchor boxes and convolutional kernels of SSD to handle the unique aspect ratios of text instances. TextBoxes++ \cite{n23} further introduces quadrangle regression to allow for multi-oriented text detection. RRD \cite{n24} extracts rotation-invariant and rotation-sensitive features for text classification and regression respectively, eliminating the incompatibility between these two tasks when detecting multi-oriented text. RRPN \cite{n25}, based on Faster R-CNN, developed Rotation Region Proposal Networks to detect multi-oriented text instances. SPCNet \cite{n26} and Mask Text Spotter \cite{n27} view text detection as an instance segmentation problem and utilize Mask R-CNN \cite{n28} for arbitrary-shaped text detection. More recent anchor-based methods, such as Boundary \cite{n44}, use Region Proposal Networks \cite{n1} and regression methods to obtain the boundary points of text instances and later rectify the detection results with differentiable operations such as Thin-Plate Splines (TPS) \cite{n46}. These structures include text recognition and enable the model to train in an end-to-end manner.
	
	Anchor-based methods normally require few post-processing steps and perform reasonably well when handling multi-oriented text. However, most of them rely on complex multiple stages and hand-crafted anchor settings, which makes these approaches overcomplicated and less effective when handling long text instances. 
		
	\begin{figure*}[!ht]
		\begin{center}
			\includegraphics[width=0.9\textwidth]{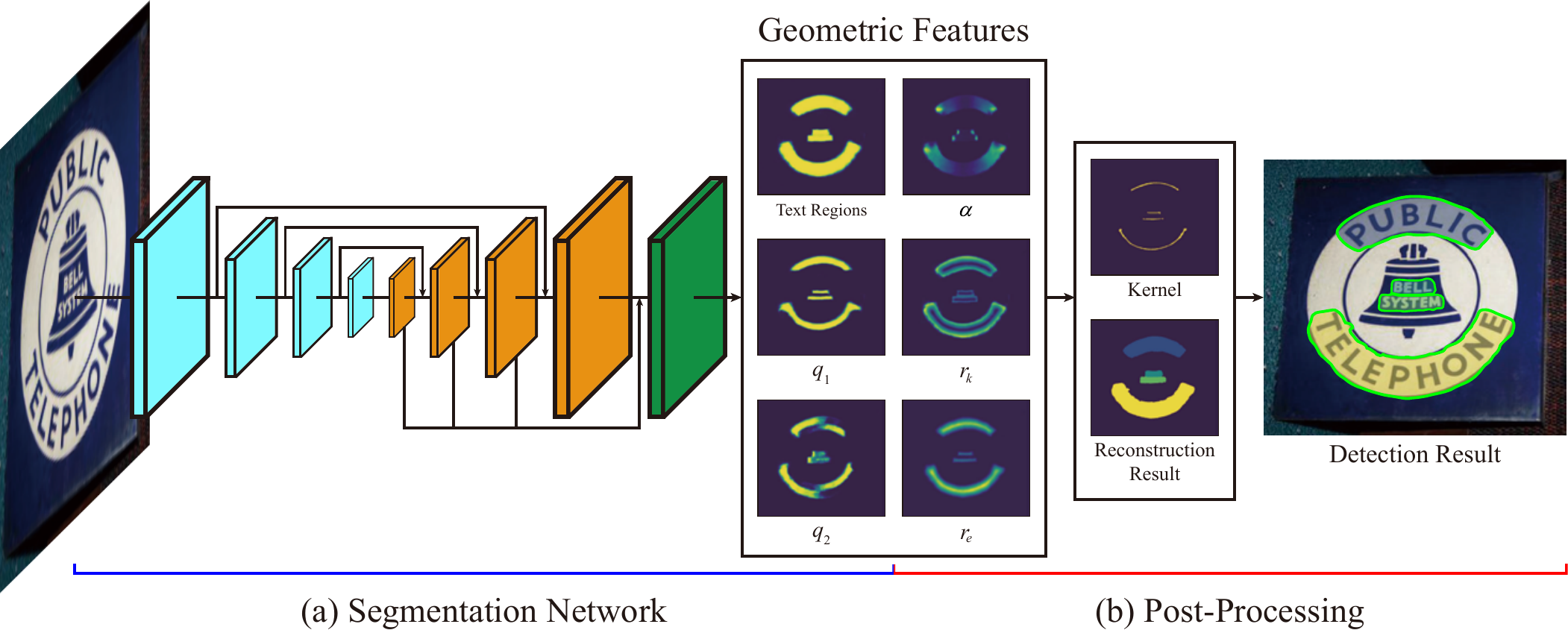}
			\caption{The inference pipeline of CTRNet. (a) The FPN-based segmentation network. The structure includes downsampling, upsampling, and the aggregation of features from different convolutional stages. For each input image, the network predicts 6 geometric features in a pixelwise manner. The produced feature map has 6 channels and is of the same width and height as the input image. (b) Illustration of the post-processing pipeline. Utilizing the predicted geometric features, a robust algorithm locates the text kernels and reconstructs the text instances. After post-processing, the detection result is obtained.}
		\end{center}
	\end{figure*}
	
	\subsection{Segmentation-Based Methods}
	Segmentation-based methods mainly view text detection as a segmentation problem and utilize Fully Convolutional Networks (FCNs) \cite{n3} or their variants. 
	
	Zhang et al. \cite{n29} first employ FCN to extract text blocks and apply MSER to detect character candidates from the text blocks. Yao et al. \cite{n30} adopt FCN to predict multiple properties of text instances, such as text regions and orientations, then implement a clustering algorithm to obtain the detection result. PixelLink \cite{n31} performs link prediction to separate adjacent text regions. EAST \cite{n9} and DeepReg \cite{n32} detect the bounding boxes of words in a per-pixel fashion without using anchors or proposal networks. Newer segmentation-based approaches, such as TextSnake \cite{n10}, PSENet \cite{n11}, and PAN \cite{n13}, detect arbitrary-shaped text instances through similar pipelines that involve text kernel segmentation and text reconstruction. Similar to anchor-based methods, there are also attempts to train segmentation-based models in an end-to-end manner. Text Perceptron \cite{n45}, for example, adopts an FPN to predict fiducial points for TPS transformations.
	
	Segmentation-based models are generally suitable for arbitrary-shaped text detection thanks to their flexibility. However, current approaches tend to endure less ideal label generation techniques and rely heavily on text kernel segmentation, which is unstable and requires deliberate tuning. In contrast, our method includes CTRs, a class of well-designed tools for label generation, and omits the need for text kernel segmentation through elaborate label design.

	\subsection{Label Generation and Post-Processing Techniques}
	Text detection methods generally adopt tailored label generation and post-processing algorithms to make full use of the geometric nature of text instances. In specific, segmentation-based models tend to develop more sophisticated label generation techniques than anchor-based models and classical text detection methods.
	
	Yin et al. \cite{new1} adopt multiple stages of post-processing steps including automatic threshold learning and machine learning-based text candidates filtering to refine the results generated by MSER. As a typical anchor-based method, RRPN \cite{n25} adopts a post-processing algorithm to compensate for the wrong detection results when handling long text instances. As for segmentation-based models, TextSnake \cite{n10} uses sophisticated label generation methods including a rule-based algorithm to extract text kernels. It later adopts a series of post-processing algorithms to reconstruct text instances and filter out false positives. PSENet uses polygon clipping for label generation. For post-processing, it utilizes an algorithm which expands the regions of the segmented text kernels to reconstruct the text instances. Moreover, according to its open-sourced official implementation, PSENet adopts a threshold-based filtering technique to filter out false positive predictions.
	
	Current label generation and post-processing techniques generally take advantage of the geometric nature of text instances. However, the label generation methods are generally rule-based, and lacks theoretical support. Also, most post-processing techniques do not utilize much of the geometric features of text instances. In contrast, the label generation process of CTRNet is based on better mathematical principles, and its post-processing pipeline fully utilizes the geometric features of text instances.
	
	\section{Methodology}
	In this section, we first introduce the inference pipeline of CTRNet. Next, we present the components of CTRNet, including the definition of a CTR, the network and label design, and the post-processing algorithm. Finally, the label generation process and the loss design are presented.
	
	\subsection{Inference Pipeline}
	The inference pipeline of the CTRNet framework is divided into two parts: a segmentation network and a post-processing algorithm (see Fig.~2).
	
	We first employ an FPN-based segmentation network to predict 6 geometric features for each pixel, as shown in Fig.~2~(a). The post-processing algorithm later utilizes the geometric features and converts them into pairs of colinear offsets from text kernels and text edges, as illustrated in Fig.~2~(b). Using the offsets, the algorithm can easily locate the text kernels and reconstruct the text instances, obtaining the detection result.
	
	For the details of the inference pipeline, Sec.~3.3.1 includes the specifics of the network structure, Sec.~3.3.2 explains the selection of the geometric features, and Sec.~3.4 introduces the process of the post-processing algorithm.
	
	\begin{figure}[!h]
		\begin{center}
			\includegraphics[width=0.45\textwidth]{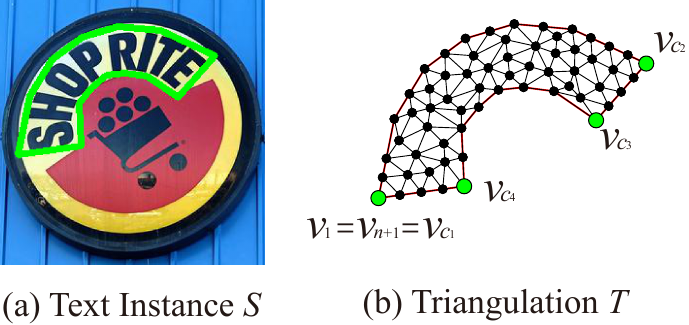}
			\caption{(a) Original text instance $S$. (b) Illustration of the triangulation and the prominent corner vertices. The prominent corners are highlighted in green. By our convention, $v_1 = v_{n+1} = v_{c_1}$. Please note that in our implementation, the triangulation has a maximum triangle area of $10^{-5}$, which it is a lot denser.}
		\end{center}
	\end{figure}

	\subsection{Conceptual Text Regions}
	In this section, we introduce Conceptual Text Regions, a class of human cognition-inspired tools aiming to describe the impression inside our mind when we conceive an arbitrary-shaped text instance. We argue that such impression ought to be a text line of rectangular shape since it is the natural form of text and corresponds to the way we read. In addition, such impression should be smooth and invertible, since we are able to conceive every pixel within the text instance.
	
	To achieve that, we develop a harmonic mapping-based rectifying method. The reason behind this choice is that harmonic mappings enable our proposed rectifying method to be smooth and bijective \cite{n37}, while other rectifying methods, such as TPS, do not ensure bijectivity \cite{n47}. To prove the necessity of using harmonic mappings instead of TPS, a comprehensive ablation study is provided in Sec.~4.3.1.

	We start with the construction of a harmonic mapping that maps an arbitrary text instance into an arbitrary rectangle. We first denote a rectangle of width $w$ and height $h$ by $R_{w,h}$. Afterwards, a harmonic mapping, denoted by $H_{w,h}$, which maps an arbitrary text instance $S$ to $R_{w,h}$ is constructed as follows:
	
	\begin{enumerate}
		\item We first define a bijective boundary mapping
		
		\begin{equation}
			b: \partial S \rightarrow \partial R_{w,h} \,,
		\end{equation}

		using the second boundary parameterization method described in Sec.~1.2.5 of \cite{n33}. The rationale behind this selection is that such method has good physical implications and ensures bijectivity \cite{n33}.
		
		\item $H_{w,h}$ is then obtained by solving the Laplace's equation
		
		\begin{equation}
			\Delta H_{w,h} \equiv 0
		\end{equation}
		
		subject to the continuous Dirichlet boundary condition
		
		\begin{equation}
			H_{w,h}|_{\partial S} = b \,.
		\end{equation}
	\end{enumerate}
	
	\begin{algorithm} [!b]
		\caption{The Calculation of $H_{w, h}$ and $H_{w, h}^{-1}$}
		\begin{algorithmic}[1]
			\REQUIRE ~\\
			Triangulation: $T$\\
			Boundary vertices: $\{v_1, ..., v_{n + 1}\}$\\
			Corner indices: $\{c_1, c_2, c_3, c_4\}$\\
			Target width and height: $\{w, h\}$\\
			\ENSURE $H_{w, h}$, $H_{w, h}^{-1}$\\
			~\\
			\COMMENT{// Obtain the bijective boundary mapping $b$.}\\
			\STATE Define $b(v_{c_1})$, $b(v_{c_2})$, $b(v_{c_3})$, and $b(v_{c_4})$ as $(0, h)$, $(w, h)$, $(w, 0)$, and $(0, 0)$, respectively.\\
			\STATE $c_5 := n+1$~~\COMMENT{// It is for convenience.}
			\FOR{$i$ in $\{1, 2, 3, 4\}$}
			\FOR{$j$ in $\{c_i, c_i + 1, ..., c_{i+1} - 1\}$}
			\STATE $b(x_j) \coloneqq \cfrac{\sum\limits_{k = c_i} ^ {j} \|v_{k + 1} - v_k\|}{\sum\limits_{k = c_i} ^ {c_{i + 1} - 1} \|v_{k + 1} - v_k\|} (b(v_{c_{i + 1}}) - b(v_{c_i})) + b(v_{c_i})$\\
			\ENDFOR\\
			\ENDFOR\\
			
			\COMMENT{// Solve the Laplace's equation on T using the FEM.}\\
			\STATE $H_{w, h}(T) \coloneqq \mathrm{FEM}(T, \Delta H_{w, h} \equiv 0, H_{w,h}|_{\partial T} = b)$\\
			\COMMENT{// Obtain $H_{w, h}$ and $H_{w, h}^{-1}$ through interpolation.}\\
			\STATE $H_{w, h}$ = $\mathrm{LinearInterpolation}(T, H_{w, h}(T))$
			\STATE $H_{w, h}^{-1}$ = $\mathrm{LinearInterpolation}(H_{w, h}(T), T)$
			
			\RETURN $H_{w, h}$, $H_{w, h}^{-1}$
		\end{algorithmic}
	\end{algorithm}

	\begin{figure}[!ht]
		\begin{center}
			\includegraphics[width=0.45\textwidth]{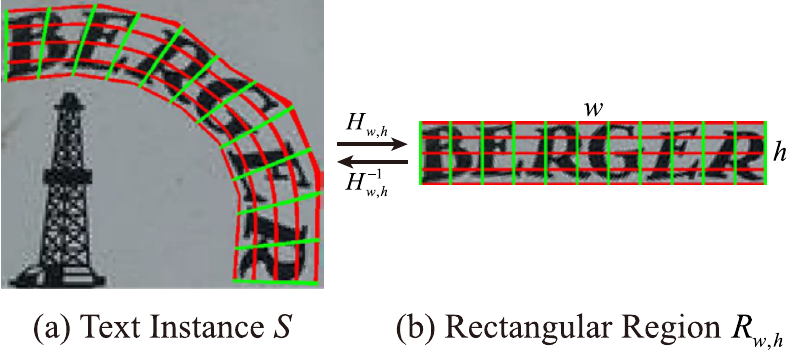}
			\caption{(a) Original text instance S. The red lines and green lines represent the horizontal lines and vertical lines mapped from $R_{w, h}$, respectively. (b) A visualization of $R_{w, h}$. The red lines depict the horizontal lines, and the green lines depict the vertical lines. Please note that choosing a different set of $w$ and $h$ for $R_{w,h}$ does not affect this mapping of horizontal and vertical lines.}
		\end{center}
	\end{figure}

	To implement the above discussed procedure, we first obtain a refined triangulation $T$ of an arbitrary-shaped text instance $S$ using the constrained Delaunay triangulation method \cite{new5} with the maximum triangle area set to $10^{-5}$ (see Fig.~3~(b)). Second, we identify the four prominent corners of the text instance according to its reading sequence, which is illustrated as the green dots in Fig.~3~(b). Please note that this piece of information is provided in most text detection datasets, including the datasets we work with. Third, we obtain the boundary vertices, $\{v_1, ..., v_{n + 1}\}$, of triangulation $T$. For convenience: 1) $v_1 = v_{n+1}$; 2) $v_1$ is the top left corner; 3) the vertices follow clockwise order. Fourth, we obtain the indices of the corner vertices, $\{c_1, c_2, c_3, c_4\}$, with $c_1 < c_2 < c_3 < c_4$. After these preparations, Algorithm 1 generates $H_{w, h}$ and $H_{w, h}^{-1}$. The Finite Element Method (FEM) \cite{new6} is used in this algorithm, and the implementation details of the FEM is provided in Sec.~4.2.
	
	Following the Radó-Kneser-Choquet theorem \cite{n34, n35, n36}, $H_{w,h}$ is bijective due to the convexity of $R_{w,h}$ \cite{n37}. With the help of $H_{w,h}$ and $H_{w,h}^{-1}$, we define the horizontal and vertical lines within $S$ by constructing $H_{1,1}^{-1}: R_{1, 1} \rightarrow S$ and afterwards mapping the horizontal and vertical lines within $R_{1, 1}$ back into $S$. Fig.~4 illustrates a visual explanation of this definition. Please note that the width and height of the rectangle $R_{w, h}$ do not affect this definition, thus, we select $R_{1, 1}$ for convenience.
		
	We may now define a CTR as $R_{w_c,h_c}$, where $w_c$ and $h_c$ are the average length of all horizontal and vertical lines within $S$, respectively. They can be calculated through the formulae:
	\begin{equation}
		w_c = \iint \limits_{R_{1,1}} \| \partial_x H_{1,1}^{-1}(x, y) \| \mathrm{d} \sigma \,,
	\end{equation}
	\begin{equation}
		h_c = \iint \limits_{R_{1,1}} \| \partial_y H_{1,1}^{-1}(x, y) \| \mathrm{d} \sigma \,.
	\end{equation}
	In our implementation, we calculate $w_c$ and $h_c$ through numerical integration.
	
	We now conclude that, for any text instance $S$, there exists a unique CTR, that is $R_{w_c,h_c}$. It inheres the bijection ${H_{w_c,h_c}: S \rightarrow R_{w_c,h_c}}$. This means that for any given $(x, y) \in S$, there is a one-to-one corresponding $(x', y') \in R_{w_c,h_c}$, and vice versa. This property will be sufficiently utilized in Sec.~3.5 for label generation.
	
	\subsection{Network and Label Design}
	In this section, we explain in detail the architecture of our segmentation network and our label design.
	
	\subsubsection{Network Design}
	The purpose of the segmentation network is to predict the geometric features which can be later used. To achieve that, we adopt the well-tested FPN-based segmentation neural network. We now introduce the details of its architecture.
	
	In a general sense, the segmentation network takes in an image of arbitrary shape and produces a feature map of the same width and height. The feature map has 6 channels, which corresponds to 3 classification results and 3 regression results.
	
	In specific, the detailed construction is demonstrated in Fig.~5. To ensure a fair comparison, we adopt the ResNet50 \cite{n6} as our backbone network, consistent to multiple recent works achieving state-of-the-art performances \cite{n11, n12, n13}. The ResNet50 structure is responsible for downsampling and is divided into 4 stages. Feature maps generated during downsampling are recorded and aggregated with their peer upsampling stages. To generate the prediction result, each one of the feature maps from the upsampling stages, illustrated in orange in Fig.~5, are upsampled and concatenated together. Finally, after a convolutional layer, a pixelwise prediction layer, and an upsampling layer, the prediction result is obtained.
	
	\begin{figure}[!ht]
		\begin{center}
			\includegraphics[width=0.45\textwidth]{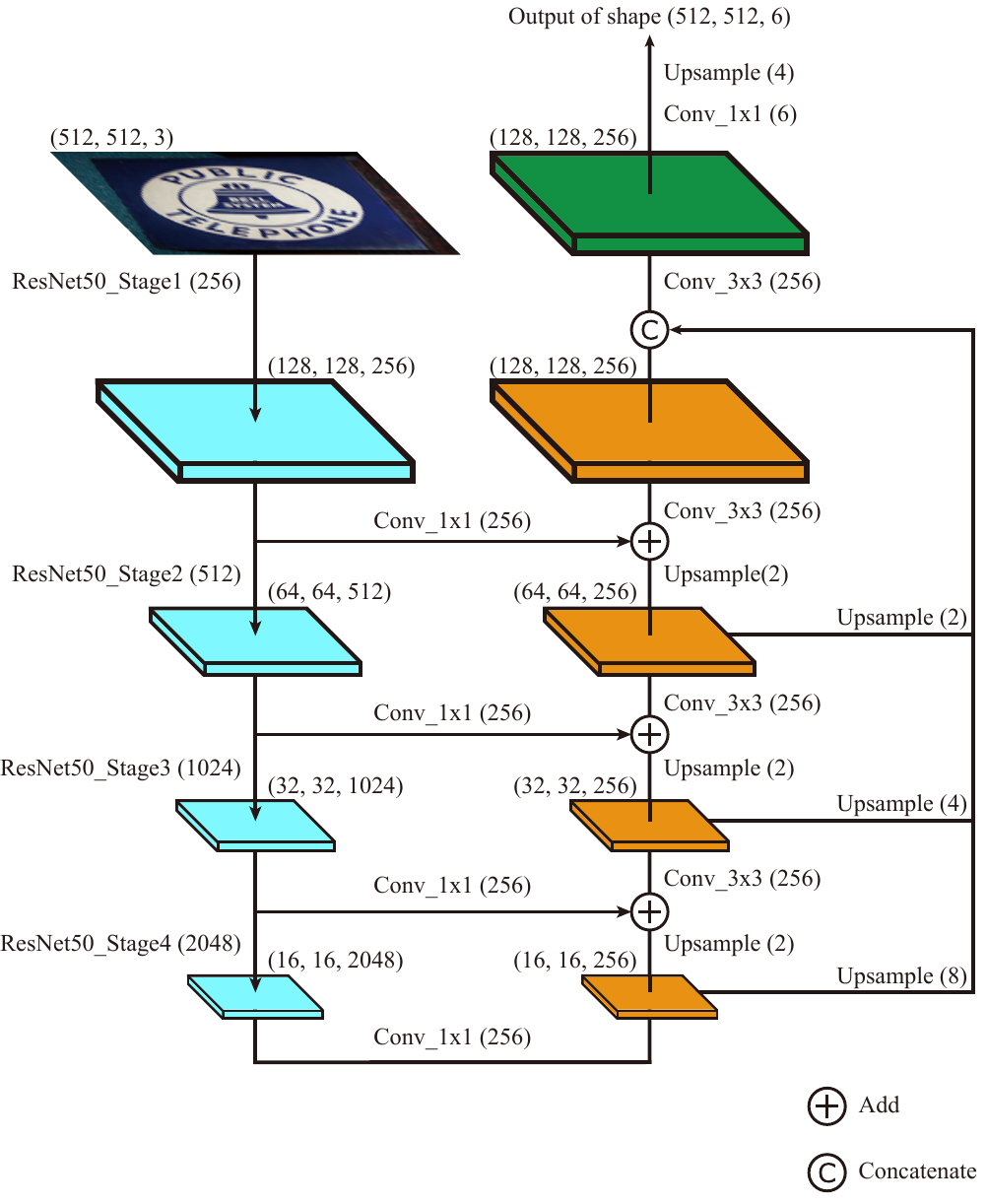}
			\caption{Illustration of the segmentation network structure. Without loss of generality, we use an image of shape $512 \times 512$ as an example, although the input shape is not constrained. The convolutional kernel sizes are represented by ``1x1'' or ``3x3''. The numbers after the convolutional stages are the output channel numbers, and the numbers after the upsampling stages are the scale factors. We use bilinear interpolation as the upsampling method to ensure smoothness.}
		\end{center}
	\end{figure}

	To train the segmentation network, label design, label generation, and a loss function are needed. They are discussed in Sec.~3.3.2, Sec.~3.5, and Sec.~3.6, respectively. Besides, hyperparameters used to train this network in our experiments are unreservedly provided in Sec.~4.2.

	\subsubsection{Label Design}
	In this section, we design the geometric features that are learnable by the segmentation network and can be later converted into the offsets required by the post-processing algorithm.
	
	For each point within a text instance, the post-processing algorithm takes in two colinear offsets and locates the corresponding text kernel and text edge. We represent each pair of those offsets with two colinear vectors ${v_k = (r_k, \theta)}$ and ${v_e = (-r_e, \theta)}$ (represented using the polar coordinate system), where $v_k$ points to the text kernel, and $v_e$ points to the text edge. Thus, for each pixel within a text instance, there is a corresponding pair of $v_k$ and $v_e$. 
	
	For our segmentation network to predict a pair of $v_k$ and $v_e$ in a pixelwise manner, we design the labels with caution to ensure they are learnable by the network. Specifically, we first use one classification head to predict the text regions. Second, we adopt two regression heads to predict radiuses $r_k$ and $r_e$. As for the angle $\theta$, instead of predicting it directly like the text regions and the radiuses, it is conventional \cite{n10} to decompose it into
	\begin{equation}
	\theta = 
	\begin{cases}
	2 \arctan{\cfrac{\sin{\theta}}{1 + \cos{\theta}}},& \mathrm{if} \cos{\theta} \neq -1\\
	\pi,& \mathrm{otherwise}\\
	\end{cases}
	\end{equation}
	and predict $\sin{\theta}$ and $\cos{\theta}$ instead. However, in our case, the feature maps of $\sin{\theta}$ and $\cos{\theta}$ exhibit drastic jump discontinuity (see Fig.~6~(c-d)). The discontinuity and rapid change make the regression heads hard to train and yield inferior results \cite{n38, n39}. Numerical experiments proving such deficiency are provided in Sec.~4.3.2. Inspired by \cite{n39}, we propose the reference angle-based encoding, which instead deconstructs $\theta$ into
	\begin{equation}
	\theta = (-1)^{q_1} \alpha + (q_1 + q_2) \pi \,,
	\end{equation}
	where $\alpha$ is the reference angle of $\theta$, and the combination of $q_1$ and $q_2$ depicts the quadrant that $\theta$ is located in. Specifically, $q_1$ is 0 if $\theta \in [0, \pi)$ and is otherwise 1. Likewise, $q_2$ is 0 if $\theta \in [\cfrac{\pi}{2}, \cfrac{3\pi}{2})$ and is otherwise 1. The motivation behind this transformation is to convert the regression target from the discontinuous $\theta$ into $\alpha$, which is generally smooth and continuous within text instances (see Fig.~6~(b)), and handle the discontinuity through classification. We subsequently predict $\alpha$, $q_1$, and $q_2$ with one regression head and two classification heads, respectively. To summarize, we adopt three classification heads to predict text regions, $q_1$, and $q_2$, and we adopt three regression heads to predict $r_k$, $r_e$, and $\alpha$.
	\begin{figure}[!t]
		\begin{center}
			\includegraphics[width=0.45\textwidth]{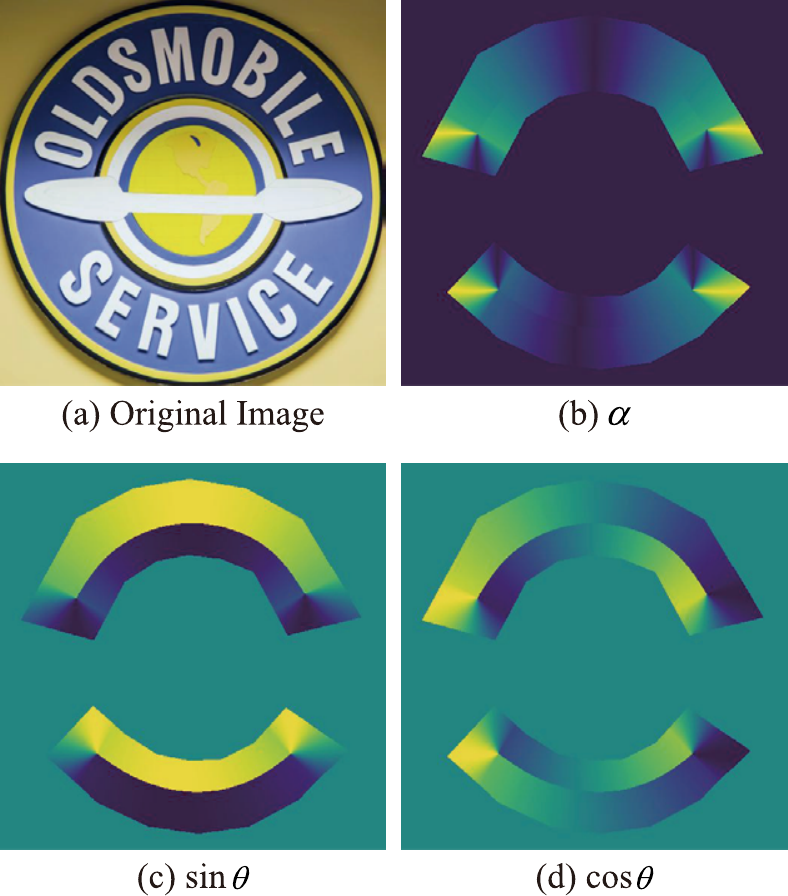}
			\caption{Illustration of the heat maps of $\alpha$, $\sin{\theta}$, and $\cos{\theta}$. It can be observed in (b) that the heat map of $\alpha$ is generally smooth and continuous, whereas in (c) and (d), drastic jump discontinuity can be observed near the text kernels.}
		\end{center}
	\end{figure}

	\subsection{Post-Processing}
	To acquire the detection result, a robust algorithm is introduced in this section to convert the geometric features predicted by the segmentation network into text predictions.
	
	\begin{figure}[!ht]
		\begin{center}
			\includegraphics[width=0.45\textwidth]{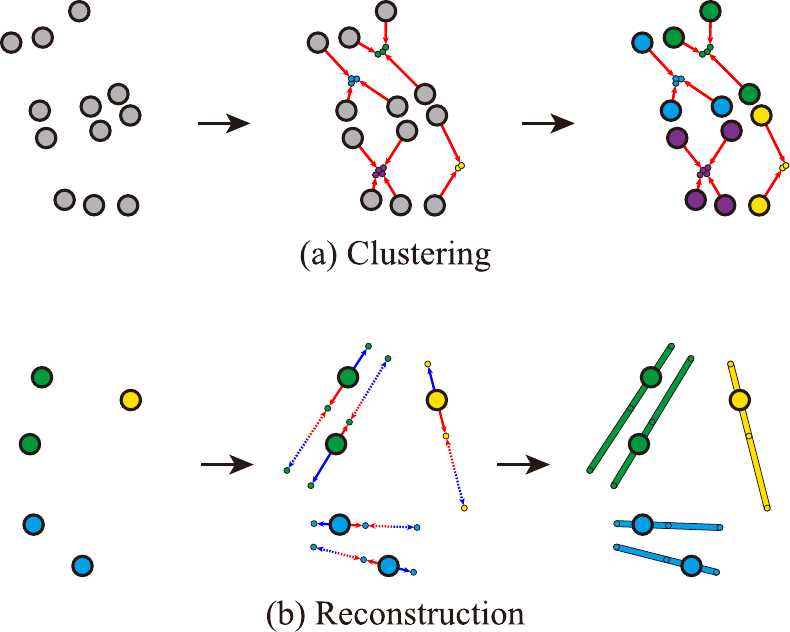}
			\caption{Illustration of the clustering and reconstruction operations: 1) The larger circles represent $p$ and the smaller circles represent either $p_k$ or $p_e$; 2) The red arrows depict $v_k$ and the blue arrows depict $v_e$; 3) The dashed arrows in (b) represent the extensions, which are of the same length as their corresponding $v_k$ or $v_e$; 4) We use distinct colors to illustrate different clusters, and being gray means unclustered.}
		\end{center}
	\end{figure}

	After prediction, the network produces 6 geometric features for each pixel, which are text regions, $r_k$, $r_e$, $\alpha$, $q_1$, and $q_2$. To begin with, we define the confidence value of a point as its confidence value for text region segmentation. As a preprocessing procedure, we first discard all points whose confidence value is lower than 0.65, and binarize $q_1$ and $q_2$ with a threshold of 0.5. For each remaining point, we then obtain its corresponding pairs of colinear vectors $v_k = (r_k, \theta)$ and $v_e = (-r_e, \theta)$, where $\theta$ is calculated through Eq.~7. Afterwards, the post-processing algorithm takes in all pairs of $v_k$ and $v_e$. It is divided into three simple operations: clustering, reconstruction, and filtering.
	
	\subsubsection{Clustering}
	As shown in Fig.~7~(a), for the points within text instances, denoted by $p$, we first locate their corresponding kernel points, defined as $p_k = p + v_k$. Next, the connected-components algorithm is applied to all kernel points $p_k$, assigning a class to each kernel point. Finally, we let all $p$ adopt the same class as their corresponding kernel points, forming distinct clusters.
	
	\subsubsection{Reconstruction}
	For each clustered point $p$, we first locate its corresponding kernel point, $p_k = p + v_k$, and edge point, $p_e = p + v_e$. Next, we further extend the line segment $\overline{p_k p_e}$ to restore the full height of the text instance, as shown in Fig. 7 (b). Finally, all points on the line segment are assigned the same class as the clustered point $p$, obtaining the reconstruction result.
	
	\subsubsection{Filtering}
	During clustering and reconstruction, rich information about the text instances are generated, which can be used to filter out false positives.
	
	Specifically, for each text instance, we first define its confidence value as the average of the confidence values within it. Second, we define its distortion as $\sigma_\alpha$, that is the standard deviation of all $\alpha$ values within the text instance. Third, we calculate its aspect ratio through formula $\cfrac{A}{4\mu_{r_k+r_e}^2}$, where $A$ represents the area of the text instance, and $\mu_{r_k+r_e}$ is the average value of $r_k + r_e$ within the text instance. Please note that this formula gives a good estimation to the aspect ratio, because
	\begin{equation}
	\cfrac{A}{4\mu_{r_k+r_e}^2} \approx \cfrac{w_c \cdot h_c}{4\mu_{r_k+r_e}^2} = \cfrac{w_c \cdot h_c}{\mu_{2r_k+2r_e}^2} \approx \cfrac{w_c \cdot h_c}{h_c^2} = \cfrac{w_c}{h_c} \,,
	\end{equation}
	where $w_c$ and $h_c$ are the width and height of the corresponding CTR. To better understand this approximation, please recall the definition of $w_c$ and $h_c$ explained in Sec.~3.2. Finally, we train a Support-Vector Machine (SVM) \cite{new4} taking the confidence values, distortion, and aspect ratios as features to filter out the false predictions.
	
	The reasons behind the choice of an SVM instead of other techniques, for example a Deep Neural Network (DNN) \cite{new7}, are twofold:
	\begin{enumerate}
		\item SVMs yield good performance with less data. In our case, the training samples are mostly identical, since most of the text instances are horizontal or alike. It results in a small amount of effective data. However, DNNs require a lot of data to avoid overfitting, while SVMs do not \cite{new3}, thus SVMs are optimal in this case.

		\item SVMs are easier to train. Unlike DNNs, SVMs generate training results quickly and deterministically \cite{new3}. Allowing us to adopt cross-validation to perform an efficient grid search for the optimal hyperparameters.
	\end{enumerate}

	Compared with the conventional false positive filtering techniques that rely on hand-crafted thresholds for confidence and area values, as observed in TextSnake \cite{n10} and PSENet \cite{n11}, our approach not only omits the need for human tuning but also leverages the more interpretable geometric information such as the distortion and aspect ratios. To validate the necessity of such geometric information, an ablation study is carried out in Sec.~4.3.3.
	
	The implementation details of the SVM are provided in Sec.~4.2. Besides, a running time analysis for the SVM is carried out in Sec.~4.5.2.
	
	\subsection{Label Generation}
	In this section, we leverage CTRs to obtain the labels designed in Sec.~3.3.2, which are text regions, $r_k$, $r_e$, $\alpha$, $q_1$, and $q_2$. For each point within a text instance, to calculate these geometric features, we need to first obtain its corresponding $v_k$ and $v_e$. These offset vectors can then be converted into the geometric features with Eq.~7.
	
	In specific, for any text instance, it takes four steps: 1) Defining the text kernel and text edge within its CTR, denoted by $C_k$ and $C_e$; 2) Defining the offset vectors within its CTR, denoted by $v_k^\prime$ and $v_e^\prime$ ; 3) Obtaining $v_k$ and $v_e$ by mapping $v_k^\prime$ and $v_e^\prime$ back into the text instance using $H_{w_c,h_c}^{-1}$; 4) Calculating the geometric features using Eq.~7. Detailed explanations are provided below.

	\subsubsection{Defining $C_k$ and $C_e$}
	For any given text instance, we first obtain its CTR, denoted by $C$, and ${C = R_{w_c,h_c}}$, where $w_c$ and $h_c$ are calculated through Eq.~4 and Eq.~5.
	
	\begin{figure}[!ht]
		\begin{center}
			\includegraphics[width=0.45\textwidth]{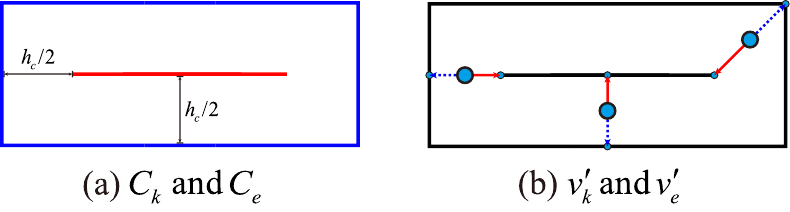}
			\caption{(a) Illustration of the text kernel and text edge defined within a CTR. The red and blue lines depict the text kernel and text edge, respectively. The black double-headed arrows indicate the lengths. (b) Illustration of the offset vectors defined within a CTR. The blue dots depict $p^\prime$, the red arrows represent $v_k^\prime$, and the blue dashed arrows represent $v_e^\prime$.}
		\end{center}
	\end{figure}

	It is trivial to define the text edge as $C_e = \partial C$. The text kernel, on the other hand, is defined as:
	\begin{equation}
		C_k = 
		\begin{cases}
		\{(x, y)|\cfrac{h_c}{2} \leq x \leq w_c - \cfrac{h_c}{2}, y = \cfrac{h_c}{2}\},& h_c < w_c\\
		\{(\cfrac{w_c}{2}, \cfrac{h_c}{2})\},& \mathrm{otherwise}\\
		\end{cases} \,.
	\end{equation}
	
	Fig.~8~(a) provides an illustration for the above definitions.
	
	\subsubsection{Defining $v_k^\prime$ and $v_e^\prime$}
	For the CTR, $C$, we use $p^\prime$ to represent a point inside it. For each $p^\prime$, it corresponds to a point in $C_k$ and a point in $C_e$. They are denoted by $p_k^\prime$ and $p_e^\prime$, respectively. For $p_k^\prime$, it is defined as the nearest point to $p^\prime$ in $C_k$. For $p_e^\prime$, on the other hand, it is defined as the intersection between $C_e$ and a ray cast from $p_k^\prime$ through $p^\prime$ (see Fig.~8~(b)). We then define the offset vectors $v_k^\prime$ and $v_e^\prime$ as:
	\begin{equation}
		v_k^\prime = p_k^\prime - p^\prime \,,
	\end{equation}	
	\begin{equation}
		v_e^\prime = p_e^\prime - p^\prime \,.
	\end{equation}
	
	An illustration for the above definitions can be found in Fig.~8~(b).

	\subsubsection{Obtaining $v_k$ and $v_e$}
	As explained in Sec.~3.2, a CTR naturally inheres a bijective harmonic mapping $H_{w_c,h_c}$ that maps any point within its corresponding text instance into itself. With the help of $H_{w_c,h_c}$, mapping $v_k^\prime$ and $v_e^\prime$ back into the original text instance is as simple as:
	\begin{equation}
		v_k = H_{w_c,h_c}^{-1}(p_k^\prime) - H_{w_c,h_c}^{-1}(p^\prime) \,,
	\end{equation}	
	\begin{equation}
		v_e = H_{w_c,h_c}^{-1}(p_e^\prime) - H_{w_c,h_c}^{-1}(p^\prime) \,.
	\end{equation}	
	
	Please note that after this mapping, $v_k$ and $v_e$ do not theoretically ensure collinearity. However, the variance is generally unnoticeable. Due to this reason, for $v_k = (r_k, \theta_k)$ and $v_e = (r_e, \theta_e)$, we use $v_e = (-r_e, \theta_k)$ to approximate the latter. The approximation is reasonable, since it hardly sacrifices any precision, and is necessary since our segmentation network is designed to handle colinear vectors. After this approximation, we use $\theta$ to represent $\theta_k$ for simplicity.
	
	\subsubsection{Obtaining the Labels}
	For the given text instance, we have obtained all of its $v_k$ and $v_e$. We now utilize Eq.~7 to convert them into the trainable geometric features required by the segmentation network.
	
	In specific, for each point within the text instance, denoted by $p$, its corresponding offsets are $v_k = (r_k, \theta)$ and $v_e = (-r_e, \theta)$. We first set the value for text regions as 1, at point $p$. Second, the features $r_k$ and $r_e$ are naturally obtained from the above expressions. Third, according to Eq.~7, we can deconstruct $\theta$ into $(-1)^{q_1} \alpha + (q_1 + q_2) \pi$, thus obtaining the values for $\alpha$, $q_1$, and $q_2$, at point $p$.
	
	To be complete, for the points that are not included in any text instance, the values for text regions are set to 0, and the values for the other geometric features are set to -1, since they are undetermined.
	
	\subsection{Loss Function}
	In this section, we introduce the loss function adopted to train the segmentation network.
	
	The loss function can be formulated as:
	\begin{equation}
		L = \lambda L_{\mathrm{text}} + (1-\lambda)(L_\alpha + L_{q_1} + L_{q_2} + L_{r_k} + L_{r_e}),
	\end{equation}
	where $\lambda$ balances the importance of $L_{\mathrm{text}}$. The binary cross-entropy loss is adopted for $L_{\mathrm{text}}$, $L_{q_1}$, and $L_{q_2}$, and the Smooth-L1 loss is selected for $L_\alpha$, $L_{r_k}$, and $L_{r_e}$, to ensure a more stable training process \cite{n1}. To overcome the data imbalance problem when segmenting text regions, we adopt OHEM 3:1 \cite{n40} for $L_{\mathrm{text}}$. Please note that all losses except for $L_{\mathrm{text}}$ are set to 0 outside text instances, since the geometric features are undefined there.
	
	\section{Experiments}
	In this section, we first conduct a comprehensive ablation study for CTRNet. Then we evaluate the proposed method on four challenging public benchmarks: CTW1500, Total-Text, MSRA-TD500, and ICDAR 2015 to compare with other state-of-the-art methods. Finally, the overall results and running time analysis are presented.
	
	\subsection{Datasets}
	\subsubsection{CTW1500}
	CTW1500 is a challenging dataset mainly focuses on long curved text instances. It consists of 1,000 training images and 500 test images, with each image containing at least 1 curved text instance.
	
	\subsubsection{Total-Text}
	Total-Text consists of word-level curved text instances. Horizontal, multi-oriented text instances are also included. It consists of 1,255 training images and 300 test images.
	
	\subsubsection{MSRA-TD500}
	MSRA-TD500 is a multi-language dataset that focuses on long multi-oriented text instances. It includes 300 training images and 200 test images.
	
	\subsubsection{ICDAR 2015}
	ICDAR 2015 focuses on incidental scenes that mainly consists of word-level multi-oriented text instances. It has 1,000 training images and 500 test images.

	\subsubsection{ICDAR 2017}
	ICDAR 2017 MLT \cite{n41} is a multi-language dataset including 9 languages representing 6 different scripts. There are 7,200 training images, 1,800 validation images and 9,000 test images in this dataset. We use this dataset only for pre-training.
	
	\subsection{Implementation Details}
	We use ResNet50 pre-trained on ImageNet \cite{n42} as our backbone. All networks are trained with a batch size of 32 on 4 GPUs and tested with a batch size of 1 on a RTX2080Ti GPU. The optimization is done by using Adam optimizer \cite{n43} with the default beta values of $(0.9, 0.999)$. 
	
	The data augmentation techniques adopted are: 1) Random rotation within range $[-10^{\circ}, 10^{\circ}]$; 2) Random resizing with ratio 0.75, 1.0, or 1.25; 3) Random cropping of size $512 \times 512$; 4) Random flipping; 5) Random color jittering.
	
	During training, the $\lambda$ for loss balancing is set to 0.67 and the negative-positive ratio of OHEM is set to 3. Following \cite{n11}, we pre-train our model on ICDAR 2017 MLT for 50K iterations with a learning rate of $1 \times 10^{-3}$. We fine-tune the pre-trained model on each benchmark dataset for 10K iterations with a learning rate of $1 \times 10^{-5}$ and compare the performance against other state-of-the-art methods. For CTW1500 dataset specifically, we also train it from scratch for 10K iterations with a learning rate of $1 \times 10^{-4}$ to convey a fair comparison against other methods that do not utilize external data.
	
	During both training and inference phase, all images are proportionally resized to ensure a suitable shorter side length (736 for ICDAR 2015 and 640 for others). 
	
	For the construction of CTRs, we use scikit-fem \cite{new2} to implement the FEM and solve the Laplace's equation. All of the partial differential equations are solved with the results recorded before training to ensure a faster training process.
	
	The SVM adopts the Radial Basis Function (RBF) \cite{new8} kernel, and its hyperparameters C and gamma are tuned through grid search within sets $\{0.1, 1, 10, 100\}$ and $\{1, 0.1, 0.001, 0.0001, 0.00001\}$, respectively. We use 5-fold cross-validation to evaluate each pair of the hyperparameters. The training data for the SVM is generated by running the segmentation model on the training set after each training epoch, so there is no test data leakage during the process.
	
	\subsection{Ablation Study}
	We conduct a comprehensive ablation study on CTW1500 and ICDAR 2015 datasets to demonstrate the effectiveness of our framework design.
	
	\begin{figure}[!ht]
		\begin{center}
			\includegraphics[width=0.45\textwidth]{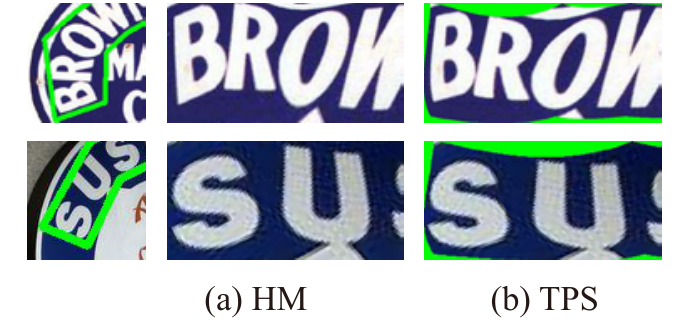}
			\caption{Comparison between the TPS method and our proposed harmonic mapping-based method. ``HM'' indicates harmonic mappings. It can be observed in (a) that the results of harmonic mappings fit the target rectangle perfectly, also they are smoother and more uniform. In contrast, the rectifying results of TPS in (b) are irregular, and they do not fill in the whole rectangle region.}
		\end{center}
	\end{figure}

	\begin{table}[!hb]
		\centering
		\caption{Experiment results under different settings. ``RAE'' indicates reference angle-based encoding, ``GI'' represents geometric information, and ``HM'' means harmonic mappings. ``P'', ``R'', and ``F'' represent precision, recall, and F-measure respectively.}
		\begin{tabular}{c@{\hskip5pt}c@{\hskip5pt}cc@{\hskip6pt}c@{\hskip6pt}cc@{\hskip6pt}c@{\hskip6pt}c}
			\hline
			\multirow{2}{*}{RAE} & \multirow{2}{*}{GI} & \multirow{2}{*}{HM} & \multicolumn{3}{c}{CTW1500} & \multicolumn{3}{c}{ICDAR 2015} \\
			\cline{4-9}
			&&& P & R & F & P & R & F\\
			\hline
			- & $\checkmark$ & $\checkmark$ & 84.9 & 72.5 & 78.2 & 86.0 & 67.4 & 75.6\\
			$\checkmark$ & - & $\checkmark$ & 86.2 & \textbf{83.6} & 84.9 & 87.3 & \textbf{83.9} & 85.6\\
			$\checkmark$ & $\checkmark$ & - & \textbf{89.7} & 80.9 & 85.1 & - & - & - \\
			$\checkmark$ & $\checkmark$ & $\checkmark$ & 88.2 & 83.3 & \textbf{85.7} & \textbf{89.5} & 83.5 & \textbf{86.4}\\
			\hline
		\end{tabular}
	\end{table}

	\subsubsection{Harmonic Mappings}
	We now explain the necessity of using harmonic mappings instead of TPS to construct CTRs.
	
	From a theoretical perspective, TPS does not promise bijectivity \cite{n47}, and there is no simple condition that can ensure a bijective TPS transformation. Thus, it is impossible for TPS to map the geometric features back into the arbitrary-shaped text instances without relying on complicated rules. In comparison, a harmonic mapping-based transformation is bijective as long as the target polygon is convex \cite{n37}, which is not only elegant but also consistent to our needs. Fig.~9 shows the quality difference between our bijective harmonic mapping-based method and the TPS method.

	From a practical perspective, although TPS performs well in many end-to-end text recognition tasks \cite{n44, n45}, it is not suitable for our CTRNet framework which requires a refined mapping. To show that, we first adopt the nearest-neighbor interpolation method to process the irregular results generated by the TPS transformation. Second, distinct negative effects on the inference pipeline caused by the TPS method can be observed in Fig.~10. Third, the numerical experiments prove our standpoints. As shown in Tab.~1, using TPS introduces a 0.6\% performance drop on CTW1500.
	
	\begin{figure}[!ht]
		\begin{center}
			\includegraphics[width=0.45\textwidth]{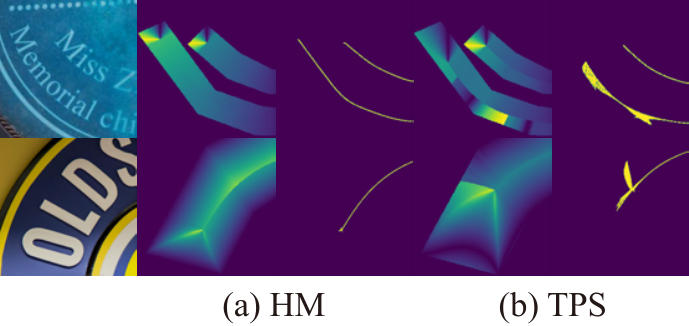}
			\caption{Comparison between the TPS method and our proposed harmonic mapping-based method. ``HM'' indicates harmonic mappings. It can be observed in (a) that the feature maps produced by harmonic mappings are smooth, and the text kernel location is accurate. However, in (b), the feature maps endure defects such as abrupt jump discontinuity, and the kernel location results are disordered.}
		\end{center}
	\end{figure}
	
	\begin{figure}[!hb]
		\begin{center}
			\includegraphics[width=0.45\textwidth]{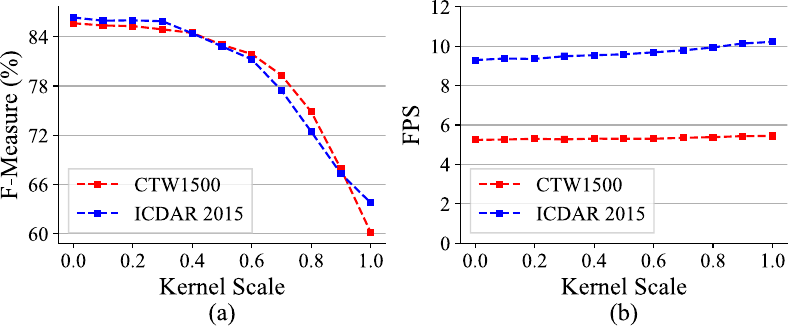}
			\caption{The influence of the kernel scale. The red lines and blue lines represent the results on CTW1500 and ICDAR 2015, respectively.}
		\end{center}
	\end{figure}

	Please note the statistical bias that a considerable portion of text instances in CTW1500 are of rectangular shapes. Considering that harmonic mappings and TPS are almost identical when handling rectangular text instances, the negative impact of TPS is a lot more significant than what the number shows. Due to the same reason, we did not perform the same experiment on ICDAR 2015. Since ICDAR 2015 only contains rectangular text instances, there is no significance in such experiment. 
	
	We also perform a comparison on the time complexity. To perform the same task of label generation for CTW1500, on average, the harmonic mapping-based method took 1.27 seconds per text instance, and the TPS method took 1.09 seconds per text instance. We conclude that both methods introduce run-time overheads, and TPS is faster. However, it does not affect the training efficiency, since label generation is only performed once before training. The labels are reused efficiently during training, with no run-time overhead.

	\subsubsection{Reference Angle-Based Encoding}
	As shown in Tab.~1, the reference angle-based encoding technique explained in Sec.~3.3.2 brings performance gains of 7.5\% and 10.8\% on CTW1500 and ICDAR 2015 respectively. Thus, we conclude that regression heads do not work well when their target feature maps exhibit abrupt jump discontinuity (see Fig.~6~(c-d)).

	\subsubsection{Geometric Information}
	As shown in Tab.~1, utilizing the additional geometric information introduced in Sec.~3.4.3, that is the distortion and aspect ratios, during the filtering process brings a 0.8\% performance gain on both CTW1500 and ICDAR 2015, proving the necessity of utilizing such information during the filtering process.
	
	\subsubsection{Kernel Scale}
	It is stated in \cite{n10} that text kernels should be granted thickness since a single-point line is prone to noise. And such statement has been practiced in various methods \cite{n10, n11, n13}. We are able to increase the kernel thickness by assigning a radius to all kernel points during post-processing. The radius is equal to $s \cdot h$, where $s$ is the kernel scale, ranging from 0 to 1, and h is the height of that text instance, given by $2 r_k + 2 r_e$. As shown in Fig.~11, in our case, the F-measure has a negative correlation with the kernel scale, while the inference speed is observed to have a positive correlation with the kernel scale. However, the gain in inference speed is minor. It is thus optimal to set the kernel scale to 0. 
	
	\subsection{Comparisons With State-of-the-Art Methods}
	We compare our proposed method with previous methods on four standard benchmarks, including two benchmarks for arbitrary-shaped text, and two benchmarks for multi-oriented text. 
	
	\begin{table}[!b]
		\centering
		\caption{Experiment results on CTW1500. ``Ext.'' indicates whether external data is used. ``P'', ``R'', and ``F'' represent precision, recall, and F-measure respectively. * indicates the results from \cite{n18}.}
		\begin{tabular}{l@{\hskip4pt}c@{\hskip4pt}ll@{\hskip4pt}l@{\hskip4pt}l}
			\hline
			Method & Ext. & Venue & P & R & F\\
			\hline
			CTPN* \cite{n8} & - & ECCV'16 & 60.4* & 53.8* & 56.9*\\
			EAST* \cite{n9} & - & CVPR'17 & 78.7* & 49.1* & 60.4*\\
			CTD+TLOC \cite{n18} & - & - & 77.4 & 69.8 & 73.4\\
			TextSnake \cite{n10} & $\checkmark$ & ECCV'18 & 67.9 & \textbf{85.3} & 75.6\\
			PSENet-1s \cite{n11} & - & CVPR'19 & 80.6 & 75.6 & 78.0\\
			CSE \cite{n48} & $\checkmark$ & CVPR'19 & 81.1 & 76.0 & 78.4\\
			PAN-640 \cite{n13} & - & ICCV'19 & 84.6 & 77.7 & 81.0\\
			PSENet-1s \cite{n11} & $\checkmark$ & CVPR'19 & 84.8 & 79.7 & 82.2\\
			Text Perceptron \cite{n45} & $\checkmark$ & AAAI'20 & \textbf{88.7} & 78.2 & 83.1\\
			DB-ResNet50 \cite{n12} & $\checkmark$ & AAAI'20 & 86.9 & 80.2 & 83.4\\
			\textbf{CTRNet} & - & - & 88.6 & 79.0 & 83.5\\
			PAN-640 \cite{n13} & $\checkmark$ & ICCV'19 & 86.4 & 81.2 & 83.7\\
			\textbf{CTRNet} & $\checkmark$ & - & 88.2 & 83.3 & \textbf{85.7}\\
			\hline
		\end{tabular}
	\end{table}	

	\subsubsection{Detecting Arbitrary-Shaped Text}
	Our CTRNet demonstrates great performance and shape robustness on two arbitrary-shaped text benchmarks: Total-Text and CTW1500.
	
	As shown in Tab.~2 and Tab.~3, our method consistently outperforms previous methods by a large margin, in terms of F-measure. Specifically, CTRNet outperforms the previous state-of-the-art methods by 2.0\% and 0.6\% on CTW1500 and Total-Text respectively, demonstrating the superiority of CTRNet when detecting arbitrary-shaped text.
	
	Moreover, when training from scratch on CTW1500, compared with other results that do not rely on external data, CTRNet outperforms the most accurate method PAN-640 by 2.5\%. Notably, it is only 0.2\% behind the previous state-of-the-art method, which is trained with external data, proving the outstanding robustness of our method.

	It is worth noting that, compared with PSENet-1s, whose network architecture is almost identical to ours, CTRNet brings huge improvements of 3.5\% on CTW1500 and 4.7\% on Total-Text. Such improvement is a solid proof for the validity of CTRs and our post-processing algorithm.

	\begin{table}[!t]
		\centering
		\caption{Experiment results on Total-Text. ``P'', ``R'', and ``F'' represent precision, recall, and F-measure respectively. * indicates the results from \cite{n10}.}
		\begin{tabular}{lllll}
			\hline
			Method & Venue & P & R & F\\
			\hline
			EAST* \cite{n9} & CVPR'17 & 50.0* & 36.2* & 42.0*\\
			TextSnake \cite{n10} & ECCV'18 & 82.7 & 74.5 & 78.4\\
			CSE \cite{n48} & CVPR'19 & 81.4 & 79.1 & 80.2\\
			PSENet-1s \cite{n11} & CVPR'19 & 84.0 & 78.0 & 80.9\\
			Text Perceptron \cite{n45} & AAAI'20 & 88.1 & 78.9 & 83.3\\
			Boundary \cite{n44} & AAAI'20 & 85.2 & \textbf{83.5} & 84.3\\
			DB-ResNet50 \cite{n12} & AAAI'20 & 87.1 & 82.5 & 84.7\\
			PAN-640 \cite{n13} & ICCV'19 & \textbf{89.3} & 81.0 & 85.0\\
			\textbf{CTRNet} & - & 88.4 & 82.9 & \textbf{85.6}\\
			\hline
		\end{tabular}
	\end{table}

	\begin{table}[!t]
		\centering
		\caption{Experiment results on MSRA-TD500. ``P'', ``R'', and ``F'' represent precision, recall, and F-measure respectively.}
		\begin{tabular}{lllll}
			\hline
			Method & Venue & P & R & F\\
			\hline
			EAST \cite{n9} & CVPR'17 & 87.3 & 67.4 & 76.1\\
			RRPN \cite{n25} & TMM'18 & 82 & 68 & 74\\
			PixelLink \cite{n31} & AAAI'18 & 83.0 & 73.2 & 77.8\\
			TextSnake \cite{n10} & ECCV'18 & 83.2 & 73.9 & 78.3\\
			RRD \cite{n24} & CVPR'18 & 87 & 73 & 79\\
			PAN \cite{n13} & ICCV'19 & 84.4 & \textbf{83.8} & 84.1\\
			DB-ResNet50 \cite{n12} & AAAI'20 & 91.5 & 79.2 & 84.9\\
			\textbf{CTRNet} & - & \textbf{92.7} & 79.1 & \textbf{85.4}\\
			\hline
		\end{tabular}
	\end{table}

	\begin{table}[!b]
		\centering
		\caption{Experiment results on ICDAR 2015. The values within parentheses indicate the height of the input image. ``P'', ``R'', and ``F'' represent precision, recall, and F-measure respectively.}
		\begin{tabular}{lllll}
			\hline
			Method & Venue & P & R & F\\
			\hline
			CTPN \cite{n8} & ECCV'16 & 74.2 & 51.6 & 60.9\\
			EAST \cite{n9} & CVPR'17 & 83.6 & 73.5 & 78.2\\
			RRPN \cite{n25} & TMM'18 & 82 & 73 & 77\\
			RRD \cite{n24} & CVPR'18 & 85.6 & 79.0 & 82.2\\
			PixelLink \cite{n31} & AAAI'18 & 82.9 & 81.7 & 82.3\\
			TextSnake \cite{n10} & ECCV'18 & 84.9 & 80.4 & 82.6\\
			PAN \cite{n13} & ICCV'19 & 84.0 & 81.9 & 82.9\\
			Boundary \cite{n44} & AAAI'20 & 88.1 & 82.2 & 85.0\\
			DB-ResNet50 (736) \cite{n12} & AAAI'20 & 88.2 & 82.7 & 85.4\\
			CSE \cite{n48} & CVPR'19 & \textbf{92.3} & 79.9 & 85.7\\
			PSENet-1s \cite{n11} & CVPR'19 & 86.9 & \textbf{84.5} & 85.7\\
			Text Perceptron \cite{n45} & AAAI'20 & 91.6 & 81.8 & 86.4\\
			DB-ResNet50 (1152) \cite{n12} & AAAI'20 & 91.8 & 83.2 & \textbf{87.3}\\
			\textbf{CTRNet} & - & 89.5 & 83.5 & 86.4\\
			\hline
		\end{tabular}
	\end{table}

	\begin{figure*}[t!]
		\begin{center}
			\includegraphics[width=1.0\textwidth]{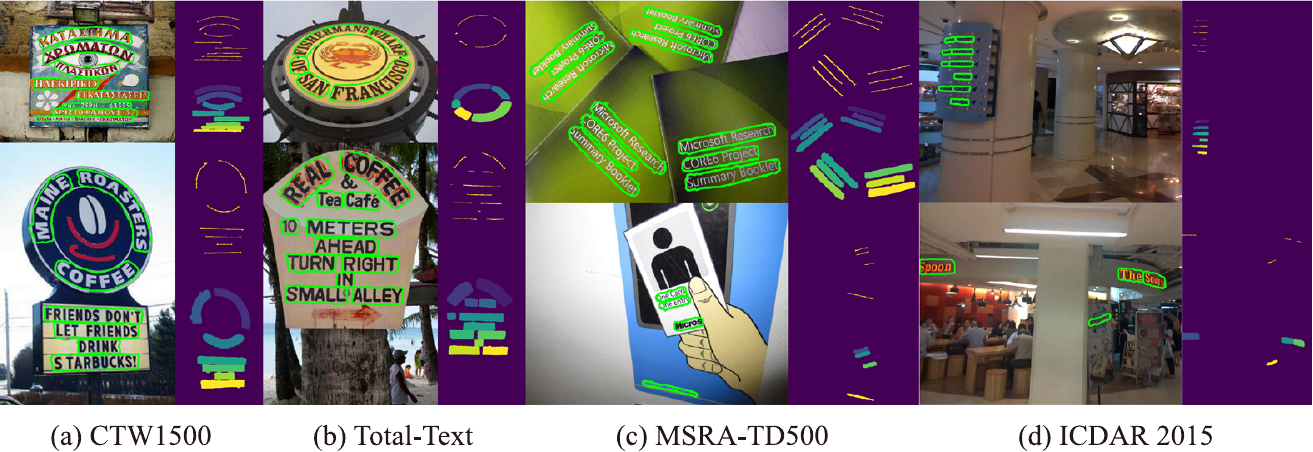}
			\caption{Qualitative results of CTRNet. Each section includes the detection result (left), kernel visualization (top right), and reconstruction result (bottom right).}
		\end{center}
	\end{figure*}

	\subsubsection{Detecting Multi-Oriented Text}
	Experiments on MSRA-TD500 and ICDAR 2015 show that CTRNet is robust when detecting multi-oriented text. 
	On MSRA-TD500, as shown in Tab.~4, CTRNet outperforms the state-of-the-art methods by 0.5\%. Specifically, it surpasses PAN and TextSnake, which relies heavily on text kernel segmentation, by solid 1.3\% and 7.1\%. Such results prove the great consistency of CTRNet and the advantage of omitting text kernel segmentation.
	
	For ICDAR 2015, it can be observed in Tab.~5 that CTRNet achieves state-of-the-art performance and is second only to DB-ResNet50 (1152). However, it is not a fair comparison since DB-ResNet50 (1152) receives input images with height 1152, which is far larger than the input images of CTRNet. When instead compared with DB-ResNet50 (736), whose input size is similar to ours, CTRNet surpasses it by 1.0\%. Further, when compared with PSENet-1s, which differs from us only in label design and post-processing, CTRNet yields an increase of 0.7\%, showing the solid improvement that our label design and post-processing algorithm bring.
	
	Please note that no specific trick, such as minimal area rectangles fitting, is implemented when handling MSRA-TD500 and ICDAR 2015, whereas some of the methods \cite{n11, n13} do. Still, the results on multi-oriented text datasets demonstrate the great consistency and robustness of CTRNet. Moreover, experiment results on MSRA-TD500 show that CTRNet is exceptionally effective when handling long text instances.
		
	\begin{table}[t!]
		\centering
		\caption{Overall results on all four benchmark datasets. ``F'' represents F-measure. ``SVM'', ``Post.'', and ``Total'' indicate the average time consumption of the SVM module, the post-processing algorithm, and the inference pipeline as a whole.}
		\begin{tabular}{lllll}
			\hline
			\multirow{2}{*}{Dataset} & \multirow{2}{*}{F} & \multicolumn{3}{c}{Time (s)}\\
			\cline{3-5}
			&& SVM & Post. & Total\\
			\hline
			CTW1500 & 85.7 & $4.83 \times 10^{-4}$ & 0.154 & 0.191\\
			Total-Text & 85.6 & $5.26 \times 10^{-4}$ & 0.092 & 0.127\\
			MSRA-TD500 & 85.4 & $1.24 \times 10^{-4}$ & 0.072 & 0.103\\
			ICDAR 2015 & 86.4 & $1.98 \times 10^{-4}$ & 0.060 & 0.108\\
			\hline
		\end{tabular}
	\end{table}
	
	\subsection{Overall Results and Running Time Analysis}
	\subsubsection{Overall Results}
	Qualitatively, it can be observed in Fig.~12 that through elaborate label design, our CTRNet generates stable and accurate predictions for text kernels, without the need of directly segmenting them. Meanwhile, CTRNet consistently produces accurate and shape-robust detection results when handling horizontal text, multi-oriented text, and arbitrary-shaped text. 
	Quantitatively, it can be observed in Tab.~6 that CTRNet consistently achieves excellence on all four of the datasets. Notably, to the best of our knowledge, CTRNet is among the first methods to achieve F-measures higher than 85.0\% on all these datasets, which proves the solid superiority of CTRNet, in terms of accuracy and consistency.
	\subsubsection{Running Time Analysis}
	We use a single RTX2080Ti GPU and the Dual Intel Xeon E5-2650 v4 @ 2.20GHz processor for running time analysis. As shown in Tab.~6, it takes an average of 0.132 seconds for CTRNet to process an image. While the SVM module poses unnoticeable time cost, the post-processing as a whole generally takes up more than 70\% of the inference time. However, it is worth noting that unlike other segmentation-based methods that implement the post-processing algorithms using the more efficient C++, such as PSENet-1s, our post-processing algorithm, for the purpose of this study, is implemented purely in Python. Nevertheless, the inference speed of CTRNet is still significantly higher than a large set of recent methods \cite{n10, n11, n48}, including PSENet-1s.
	
	To carry out a fair comparison, we implement PSENet-1s using its open-sourced official implementation, under our hardware environment. According to our tests on CTW1500, its average running time for each picture is 0.334 seconds. It is slower than CTRNet, which takes 0.191 seconds under the same experimental configuration. This shows that CTRNet has a competitive running time performance.

	\section{Conclusion}
	In this paper, we propose an effective framework to detect arbitrary-shaped text with outstanding accuracy and stability. We first introduce Conceptual Text Regions, a class of cognition-inspired tools for elaborate label generation. Further, we propose a well-designed inference pipeline to predict text instances using the geometric features generated by CTRs. The fact that CTRs allow for sophisticated label design and that the inference pipeline omits the need for text kernel segmentation establishes CTRNet as a robust and accurate arbitrary-shaped text detector. Achieving F-measures greater than 85.0\% on all four of CTW1500, Total-Text, MSRA-TD500, and ICDAR 2015, CTRNet demonstrates superior advantages in terms of accuracy and consistency when compared with previous state-of-the-art text detectors.
	
	For future study, we consider it meaningful to investigate the possibility of establishing the CTR method as a universal text label generation technique. It is also of interest to utilize the potential of CTRs and evolve CTRNet into an end-to-end trainable text recognition framework. Last but not least, we will seek to utilize deep reinforcement learning and multi-task learning algorithms \cite{n49, n50, n51, n52} to optimize the generalization capability of such text recognition framework.


\newpage
	
	\begin{thebibliography}{99}
		\bibitem{n1}
		S. Ren, K. He, R. Girshick, J. Sun,
		Faster r-cnn: Towards real-time object detection with region proposal networks,
		IEEE Trans. Pattern Anal. Mach. Intell.
		39 (2017) 1137-1149.
		\url{https://doi.org/10.1109/TPAMI.2016.2577031}.
		
		\bibitem{n2}
		W. Liu, D. Anguelov, D. Erhan, C. Szegedy, S. Reed, C.Y. Fu, A.C. Berg,
		Ssd: Single shot multibox detector,
		in: Proc. Eur. Conf. Comput. Vis.,
		2016: pp. 21-37.
		\url{https://doi.org/10.1007/978-3-319-46448-0_2}.
		
		\bibitem{n3}
		J. Long, E. Shelhamer, T. Darrell,
		Fully convolutional networks for semantic segmentation,
		in: Proc. IEEE Conf. Comput. Vis. Pattern Recognit.,
		2015: pp. 3431-3440.
		\url{https://doi.org/10.1109/CVPR.2015.7298965}.
		
		\bibitem{n4}
		O. Ronneberger, P. Fischer, T. Brox,
		U-net: Convolutional networks for biomedical image segmentation,
		in: Proc. Int. Conf. Med. Image Comput. Comput. Interv.,
		2015: pp. 234-241.
		\url{https://doi.org/10.1007/978-3-319-24574-4_28}.
		
		\bibitem{n5}
		Q. Ning, J. Zhu, C. Chen,
		Very fast semantic image segmentation using hierarchical dilation and feature refining,
		Cognit. Comput.
		10 (2018) 62-72.
		\url{https://doi.org/10.1007/s12559-017-9530-0}.
		
		\bibitem{n6}
		K. He, X. Zhang, S. Ren, J. Sun,
		Deep residual learning for image recognition,
		in: Proc. IEEE Conf. Comput. Vis. Pattern Recognit.,
		2016: pp. 770-778.
		\url{https://doi.org/10.1109/CVPR.2016.90}.
		
		\bibitem{n7}
		G. Huang, Z. Liu, L.V.D. Maaten, K.Q. Weinberger,
		Densely connected convolutional networks,
		in: Proc. IEEE Conf. Comput. Vis. Pattern Recognit.,
		2017: pp. 2261-2269.
		\url{https://doi.org/10.1109/CVPR.2017.243}.
		
		\bibitem{n8}
		Z. Tian, W. Huang, T. He, P. He, Y. Qiao, 
		Detecting text in natural image with connectionist text proposal network,
		in: Proc. Eur. Conf. Comput. Vis.,
		2016: pp. 56-72.
		\url{https://doi.org/10.1007/978-3-319-46484-8_4}.
		
		\bibitem{n9}
		X. Zhou, C. Yao, H. Wen, Y. Wang, S. Zhou, W. He, J. Liang,
		East: An efficient and accurate scene text detector,
		in: Proc. IEEE Conf. Comput. Vis. Pattern Recognit.,
		2017: pp. 2642-2651.
		\url{https://doi.org/10.1109/CVPR.2017.283}.
		
		\bibitem{n10}
		S. Long, J. Ruan, W. Zhang, X. He, W. Wu, C. Yao,
		Textsnake: A flexible representation for detecting text of arbitrary shapes,
		in: Proc. Eur. Conf. Comput. Vis.,
		2018: pp. 19-35.
		\url{https://doi.org/10.1007/978-3-030-01216-8_2}.
		
		\bibitem{n11}
		W. Wang, E. Xie, X. Li, W. Hou, T. Lu, G. Yu, S. Shao,
		Shape robust text detection with progressive scale expansion network,
		in: Proc. IEEE Conf. Comput. Vis. Pattern Recognit.,
		2019: pp. 9328-9337.
		\url{https://doi.org/10.1109/CVPR.2019.00956}.
		
		\bibitem{n12}
		M. Liao, Z. Wan, C. Yao, K. Chen, X. Bai,
		Real-time scene text detection with differentiable binarization,
		in: Proc. AAAI Conf. Artif. Intell.,
		2020: pp. 11474-11481.
		\url{https://doi.org/10.1609/aaai.v34i07.6812}.
		
		\bibitem{n13}
		W. Wang, E. Xie, X. Song, Y. Zang, W. Wang, T. Lu, G. Yu, C. Shen,
		Efficient and accurate arbitrary-shaped text detection with pixel aggregation network,
		in: Proc. IEEE Int. Conf. Comput. Vis.,
		2019: pp. 8440-8449.
		\url{https://arxiv.org/abs/1908.05900}.
		
		\bibitem{n14}
		T.Y. Lin, P. Dollár, R. Girshick, K. He, B. Hariharan, S. Belongie,
		Feature pyramid networks for object detection,
		in: Proc. IEEE Conf. Comput. Vis. Pattern Recognit.,
		2017: pp. 936-944.
		\url{https://doi.org/10.1109/CVPR.2017.106}.
		
		\bibitem{n15}
		F. Wang, X. Wang, J. Tang, B. Luo, C. Li,
		Vtann: Visual tracking with attentive adversarial network,
		Cognit. Comput.
		(2020). \url{https://doi.org/10.1007/s12559-020-09727-3}.
		
		\bibitem{n16}
		B. Farhadinia,
		A cognitively inspired knowledge-based decision-making methodology employing intuitionistic fuzzy sets,
		Cognit. Comput.
		12 (2020) 667-678.
		\url{https://doi.org/10.1007/s12559-019-09702-7}.
		
		\bibitem{n17}
		X. Bi, X. Zhao, H. Huang, D. Chen, Y. Ma,
		Functional brain network classification for alzheimer's disease detection with deep features and extreme learning machine,
		Cognit. Comput.
		12 (2020) 513-527.
		\url{https://doi.org/10.1007/s12559-019-09688-2}.
		
		\bibitem{n18}
		Y. Liu, L. Jin, S. Zhang, S. Zhang,
		Detecting curve text in the wild: New dataset and new solution,
		2017.
		\url{https://arxiv.org/abs/1712.02170}.
		
		\bibitem{n19}
		C.K. Ch'ng, C.S. Chan,
		Total-text: A comprehensive dataset for scene text detection and recognition,
		in: Proc. IAPR Int. Conf. Doc. Anal. Recognit.,
		2017: pp. 935-942.
		\url{https://doi.org/10.1109/ICDAR.2017.157}.
		
		\bibitem{n20}
		C. Yao, X. Bai, W. Liu, Y. Ma, Z. Tu,
		Detecting texts of arbitrary orientations in natural images,
		in: Proc. IEEE Conf. Comput. Vis. Pattern Recognit.,
		2012: pp. 1083-1090.
		\url{https://doi.org/10.1109/CVPR.2012.6247787}.
		
		\bibitem{n21}
		D. Karatzas, L. Gomez-Bigorda, A. Nicolaou, S. Ghosh, A. Bagdanov, M. Iwamura, J. Matas, L. Neumann, V.R. Chandrasekhar, S. Lu, F. Shafait, S. Uchida, E. Valveny,
		Icdar 2015 competition on robust reading,
		in: Proc. IAPR Int. Conf. Doc. Anal. Recognit.,
		2015: pp. 1156-1160.
		\url{https://doi.org/10.1109/ICDAR.2015.7333942}.
		
		\bibitem{n22}
		M. Liao, B. Shi, X. Bai, X. Wang, W. Liu,
		Textboxes: A fast text detector with a single deep neural network,
		in: Proc. AAAI Conf. Artif. Intell.,
		2017: pp. 4161-4167.
		\url{https://arxiv.org/abs/1611.06779}.
		
		\bibitem{n23}
		M. Liao, B. Shi, X. Bai,
		Textboxes++: A single-shot oriented scene text detector,
		IEEE Trans. Image Process.
		27 (2018) 3676-3690.
		\url{https://doi.org/10.1109/TIP.2018.2825107}.
		
		\bibitem{n24}
		M. Liao, Z. Zhu, B. Shi, G.S. Xia, X. Bai,
		Rotation-sensitive regression for oriented scene text detection,
		in: Proc. IEEE Conf. Comput. Vis. Pattern Recognit.,
		2018: pp. 5909-5918.
		\url{https://doi.org/10.1109/CVPR.2018.00619}.
		
		\bibitem{n25}
		J. Ma, W. Shao, H. Ye, L. Wang, H. Wang, Y. Zheng, X. Xue,
		Arbitrary-oriented scene text detection via rotation proposals,
		IEEE Trans. Multimed.
		20 (2018) 3111-3122.
		\url{https://doi.org/10.1109/TMM.2018.2818020}.
		
		\bibitem{n26}
		E. Xie, Y. Zang, S. Shao, G. Yu, C. Yao, G. Li,
		Scene text detection with supervised pyramid context network,
		in: Proc. AAAI Conf. Artif. Intell.,
		2019: pp. 9038-9045.
		\url{https://doi.org/10.1609/aaai.v33i01.33019038}.
		
		\bibitem{n27}
		M. Liao, P. Lyu, M. He, C. Yao, W. Wu, X. Bai,
		Mask textspotter: An end-to-end trainable neural network for spotting text with arbitrary shapes,
		IEEE Trans. Pattern Anal. Mach. Intell.
		43 (2021) 532-548.
		\url{https://doi.org/10.1109/tpami.2019.2937086}.
		
		\bibitem{n28}
		K. He, G. Gkioxari, P. Dollár, R. Girshick,
		Mask r-cnn,
		in: Proc. IEEE Int. Conf. Comput. Vis.,
		2017: pp. 2980-2988.
		\url{https://doi.org/10.1109/ICCV.2017.322}.
		
		\bibitem{n44}
		H. Wang, P. Lu, H. Zhang, M. Yang, X. Bai, Y. Xu, M. He, Y. Wang, W. Liu,
		All you need is boundary: Toward arbitrary-shaped text spotting,
		in: Proc. AAAI Conf. Artif. Intell.,
		2020: pp. 12160-12167.
		\url{https://doi.org/10.1609/aaai.v34i07.6896}.
		
		\bibitem{n46}
		F.L. Bookstein,
		Principal warps: Thin-plate splines and the decomposition of deformations,
		IEEE Trans. Pattern Anal. Mach. Intell.
		11 (1989) 567-585.
		\url{https://doi.org/10.1109/34.24792}.
		
		\bibitem{n29}
		Z. Zhang, C. Zhang, W. Shen, C. Yao, W. Liu, X. Bai,
		Multi-oriented text detection with fully convolutional networks,
		in: Proc. IEEE Conf. Comput. Vis. Pattern Recognit.,
		2016: pp. 4159-4167.
		\url{https://doi.org/10.1109/CVPR.2016.451}.
		
		\bibitem{n30}
		C. Yao, X. Bai, N. Sang, X. Zhou, S. Zhou, Z. Cao,
		Scene text detection via holistic, multi-channel prediction,
		2016.
		\url{https://arxiv.org/abs/1606.09002}.
		
		\bibitem{n31}
		D. Deng, H. Liu, X. Li, D. Cai,
		Pixellink: Detecting scene text via instance segmentation,
		in: Proc. AAAI Conf. Artif. Intell.,
		2018: pp. 6773-6780.
		\url{https://arxiv.org/abs/1801.01315}.
		
		\bibitem{n32}
		W. He, X.Y. Zhang, F. Yin, C.L. Liu,
		Deep direct regression for multi-oriented scene text detection,
		in: Proc. IEEE Int. Conf. Comput. Vis.,
		2017: pp. 745-753.
		\url{https://doi.org/10.1109/ICCV.2017.87}.
		
		\bibitem{n45}
		L. Qiao, S. Tang, Z. Cheng, Y. Xu, Y. Niu, S. Pu, F. Wu,
		Text perceptron: Towards end-to-end arbitrary-shaped text spotting,
		in: Proc. AAAI Conf. Artif. Intell.,
		2020: pp. 11899-11907.
		\url{https://doi.org/10.1609/aaai.v34i07.6864}.
		
		\bibitem{new1}
		X.C. Yin, X. Yin, K. Huang, H.W. Hao,
		Robust text detection in natural scene images,
		IEEE Trans. Pattern Anal. Mach. Intell.
		36 (2014) 970-983.
		\url{https://doi.org/10.1109/TPAMI.2013.182}.

		\bibitem{n37}
		T. Schneider, K. Hormann,
		Smooth bijective maps between arbitrary planar polygons,
		Comput. Aided Geom. Des.
		35-36 (2015) 243-254.
		\url{https://doi.org/10.1016/j.cagd.2015.03.010}.
		
		\bibitem{n47}
		A.P. Erikson, K. Åström,
		On the bijectivity of thin-plate splines,
		Anal. Sci. Eng. Beyond.
		6 (2012) 93-141.
		\url{https://doi.org/10.1007/978-3-642-20236-0_5}.

		\bibitem{n33}
		K. Hormann,
		Theory and applications of parameterizing triangulations,
		2001.

		\bibitem{new5}
		L.P. Chew,
		Constrained delaunay triangulations,
		Algorithmica.
		4 (1989) 97-108.
		\url{https://doi.org/10.1007/BF01553881}.
		
		\bibitem{new6}
		O.C. Zienkiewicz, R.L. Taylor, J.Z. Zhu,
		The finite element method: Its basis and fundamentals,
		Elsevier.
		2005.
		
		\bibitem{n34}
		T. Radó,
		Aufgabe 41,
		Jahresbericht Der Dtsch. Math.
		35 (1926) 49-49.
		
		\bibitem{n35}
		H. Kneser,
		Lösung der aufgabe 41,
		Jahresbericht Der Dtsch. Math.
		35 (1926) 123-124.
		
		\bibitem{n36}
		G. Choquet,
		Sur un type de transformation analytique généralisant la représentation conforme et définie au moyen de fonctions harmoniques,
		Bull. Des Sci. Mathématiques.
		69 (1945) 156-165.
		
		\bibitem{n38}
		R.R. Selmic, F.L. Lewis,
		Neural-network approximation of piecewise continuous functions: Application to friction compensation,
		IEEE Trans. Neural Networks.
		13 (2002) 745-751.
		\url{https://doi.org/10.1109/TNN.2002.1000141}.
		
		\bibitem{n39}
		B. Llanas, S. Lantarón, F.J. Sáinz,
		Constructive approximation of discontinuous functions by neural networks,
		Neural Process. Lett.
		27 (2008) 209-226.
		\url{https://doi.org/10.1007/s11063-007-9070-9}.
		
		\bibitem{new4}
		W.S. Noble,
		What is a support vector machine?,
		Nat. Biotechnol.
		24 (2006) 1565-1567.
		\url{https://doi.org/10.1038/nbt1206-1565}.
		
		\bibitem{new7}
		P. Sharma, A. Singh,
		Era of deep neural networks: A review,
		in: Proc. Int. Conf. Comput. Commun. Netw. Technol.,
		2017: pp. 1-5.
		/url{https://doi.org/10.1109/ICCCNT.2017.8203938}.
		
		\bibitem{new3}
		K. Pasupa, W. Sunhem,
		A comparison between shallow and deep architecture classifiers on small dataset,
		in: Proc. Int. Conf. Inf. Technol. Electr. Eng.,
		2016: pp. 1-6.
		\url{https://doi.org/10.1109/ICITEED.2016.7863293}.
		
		\bibitem{n40}
		A. Shrivastava, A. Gupta, R. Girshick,
		Training region-based object detectors with online hard example mining,
		in: Proc. IEEE Conf. Comput. Vis. Pattern Recognit.,
		2016: pp. 761-769.
		\url{https://doi.org/10.1109/CVPR.2016.89}.
		
		\bibitem{n41}
		N. Nayef, F. Yin, I. Bizid, H. Choi, Y. Feng, D. Karatzas, Z. Luo, U. Pal, C. Rigaud, J. Chazalon, W. Khlif, M.M. Luqman, J.C. Burie, C.L. Liu, J.M. Ogier,
		Icdar2017 robust reading challenge on multi-lingual scene text detection and script identification - rrc-mlt,
		in: Proc. IAPR Int. Conf. Doc. Anal. Recognit.,
		2017: pp. 1454-1459.
		\url{https://doi.org/10.1109/ICDAR.2017.237}.
		
		\bibitem{n42}
		J. Deng, W. Dong, R. Socher, L.J. Li, K. Li, L. Fei-Fei,
		Imagenet: A large-scale hierarchical image database,
		in: Proc. IEEE Conf. Comput. Vis. Pattern Recognit.,
		2009: pp. 248-255.
		\url{https://doi.org/10.1109/CVPR.2009.5206848}.
		
		\bibitem{n43}
		D.P. Kingma, J.L. Ba,
		Adam: A method for stochastic optimization,
		in: Proc. Int. Conf. Learn. Represent.,
		2015.
		\url{https://arxiv.org/abs/1412.6980}.
		
		\bibitem{new2}
		T. Gustafsson, G.D. McBain,
		Scikit-fem: A python package for finite element assembly,
		J. Open Source Softw.
		5 (2020) 2369-2369.
		\url{https://doi.org/10.21105/joss.02369}.	
		
		\bibitem{new8}
		B. Scholkopf, K.K. Sung, C.J.C. Burges, F. Girosi, P. Niyogi, T. Poggio, V. Vapnik,
		Comparing support vector machines with gaussian kernels to radial basis function classifiers,
		IEEE Trans. Signal Process.
		45 (1997) 2758-2765.
		\url{https://doi.org/10.1109/78.650102}.
		
		\bibitem{n48}
		Z. Liu, G. Lin, S. Yang, F. Liu, W. Lin, W.L. Goh,
		Towards robust curve text detection with conditional spatial expansion,
		in: Proc. IEEE Conf. Comput. Vis. Pattern Recognit.,
		2019: pp. 7261-7270.
		\url{https://doi.org/10.1109/CVPR.2019.00744}.
		
		\bibitem{n49}
		N. Chouikhi, B. Ammar, A. Hussain, A.M. Alimi,
		Bi-level multi-objective evolution of a multi-layered echo-state network autoencoder for data representations,
		Neurocomputing.
		341 (2019) 195-211.
		\url{https://doi.org/10.1016/j.neucom.2019.03.012}.
		
		\bibitem{n50}
		C. Ieracitano, N. Mammone, A. Bramanti, A. Hussain, F.C. Morabito,
		A convolutional neural network approach for classification of dementia stages based on 2d-spectral representation of eeg recordings,
		Neurocomputing.
		323 (2019) 96-107.
		\url{https://doi.org/10.1016/j.neucom.2018.09.071}.
		
		\bibitem{n51}
		M. Mahmud, M.S. Kaiser, A. Hussain, S. Vassanelli,
		Applications of deep learning and reinforcement learning to biological data,
		IEEE Trans. Neural Networks Learn. Syst.
		29 (2018) 2063-2079.
		\url{https://doi.org/10.1109/TNNLS.2018.2790388}.
		
		\bibitem{n52}
		F. Xiong, B. Sun, X. Yang, H. Qiao, K. Huang, A. Hussain, Z. Liu,
		Guided policy search for sequential multitask learning,
		IEEE Trans. Syst. Man Cybern. Syst.
		49 (2019) 216-226.
		\url{https://doi.org/10.1109/TSMC.2018.2800040}.
	\end{thebibliography}
\end{document}